\documentclass[11pt,letterpaper]{article}
\usepackage{emnlp2016}
\usepackage{times}
\usepackage{latexsym}

\usepackage{array}
\usepackage{color}
\usepackage{upgreek,amssymb,amsmath}
\usepackage{colortbl}
\definecolor{grey}{rgb}{0.8, 0.8, 0.84}
\definecolor{grey2}{rgb}{0.92, 0.92, 0.97}

\usepackage{hhline}

\usepackage{scrextend}
\usepackage{hyperref}

\usepackage{graphicx}
\usepackage{amssymb,amsmath,bm}
\usepackage{textcomp}

\newcolumntype{X}[1]{>{\centering\arraybackslash}p{#1}}
\newcolumntype{A}{>{\centering\arraybackslash}p{0.4cm}}
\newcolumntype{B}{>{\centering\arraybackslash}p{0.5cm}}
\newcolumntype{C}{>{\centering\arraybackslash}p{0.4cm}}
\newcolumntype{D}{>{\centering\arraybackslash}p{4.59cm}}

\emnlpfinalcopy

\def\emnlppaperid{640}

\title{Long-Short Range Context Neural Networks for Language Modeling}

\author{Youssef Oualil\textsuperscript{1,2} \and Mittul Singh\textsuperscript{1,3} \and 
Clayton Greenberg\textsuperscript{1,2,3} \and Dietrich Klakow\textsuperscript{1,2,3}  \\
\textsuperscript{1}Spoken Language Systems (LSV) \\
\textsuperscript{2}Collaborative Research Center on Information Density and Linguistic Encoding  \\
\textsuperscript{3}Graduate School of Computer Science \\
Saarland University, Saarbr\"{u}cken, Germany \\
\tt \{firstname.lastname\}@lsv.uni-saarland.de
}

\date{}

\begin{document}

\maketitle

\begin{abstract}
The goal of language modeling techniques is to capture the statistical and structural properties of natural 
languages from training corpora. This task typically involves the learning of short range dependencies, 
which generally model the syntactic properties of a language and/or long range dependencies, which are semantic in nature. 
We propose in this paper a new multi-span architecture, which separately models the short and long 
context information while it dynamically merges them to perform the language modeling task. This is 
done through a novel recurrent Long-Short Range Context (LSRC) network, which 
explicitly models the local (short) and global (long) context using two separate hidden states 
that evolve in time. This new architecture is an adaptation of the Long-Short Term Memory network 
(LSTM) to take into account the linguistic properties. Extensive experiments conducted on the Penn Treebank (PTB) 
and the Large Text Compression Benchmark (LTCB) corpus showed a significant
reduction of the perplexity when compared to state-of-the-art language modeling techniques.
\end{abstract}

\section{Introduction}
\label{sec:intro}

A high quality Language Model (LM) is considered to be an integral component of many systems for 
speech and language technology applications, such as machine translation~\cite{Brown1990}, speech recognition~\cite{Katz1987}, etc.  
The goal of an LM is to identify and predict probable sequences of predefined linguistic units, 
which are typically words. These predictions are typically guided by the semantic and syntactic properties encoded by the LM.

In order to capture these properties, classical LMs were typically developed as fixed (short) 
context techniques such as, the word count-based methods~\cite{Rosenfeld2000,KN1995}, 
commonly known as $N$-gram language models, as well as the Feedforward Neural Networks 
(FFNN)~\cite{Bengio2003}, which were introduced as an alternative to overcome the exponential 
growth of parameters required for larger context sizes in $N$-gram models. 

In order to overcome the short context constraint and capture long range dependencies known to 
be present in language,~\newcite{Bellegarda:1998} proposed to use Latent Semantic Analysis (LSA) 
to capture the global context, and then combine it with the standard $N$-gram models, which capture 
the local context. In a similar but more recent approach,~\newcite{Mikolov2012} showed that Recurrent 
Neural Network (RNN)-based LM performance can be significantly improved using an additional global 
topic information obtained using Latent Dirichlet Allocation (LDA). In fact,    
 although recurrent architectures theoretically allow the context to indefinitely cycle in the 
network,~\newcite{HaiSon2012} have shown that, in practice, this information changes quickly in the 
classical RNN~\cite{Mikolov2010} structure, and that it is experimentally equivalent to an 8-gram FFNN. 
Another alternative to model linguistic dependencies, Long-Short Term Memory (LSTM)~\cite{Sundermeyer12}, 
addresses some learning issues from the original RNN by controlling the longevity of context information 
in the network. This architecture, however, does not particularly model long/short context but rather 
uses a single state to model the global linguistic context. 

Motivated by the works in~\cite{Bellegarda:1998,Mikolov2012}, this paper proposes a novel neural 
architecture which explicitly models 1) the local (short) context information, generally syntactic, 
as well as 2) the global (long) context, which is semantic in nature, using two separate recurrent 
hidden states. These states evolve in parallel within a long-short range context network. In doing so, 
the proposed architecture is particularly adapted to model natural languages that manifest 
local-global context information in their linguistic properties.   

We proceed as follows.  Section~\ref{sec:SLCLM} presents a brief overview of short vs long range context language modeling techniques. 
Section~\ref{sec:MSLM} introduces the novel architecture, Long-Short Range Context (LSRC), which explicitly models these two dependencies. 
Then, Section~\ref{sec:EXP} evaluates the proposed network in comparison to different state-of-the-art language models on the PTB and the LTCB corpus. 
Finally, we conclude in Section~\ref{sec:CC}.

\section{Short vs Long Context Language Models}
\label{sec:SLCLM}

The goal of a language model is to estimate the probability distribution $p(w_1^T)$ of word sequences $w_1^T = w_1,\cdots,w_T$.
Using the chain rule, this distribution can be expressed as
\vspace{-2mm}
\begin{equation}
  \label{eq:prob}
\displaystyle{ p(w_1^T) = \prod_{t=1}^T{p(w_t|w_1^{t-1})} }
\vspace{-2mm}
\end{equation}
This probability is generally approximated under different simplifying assumptions, which are typically derived based on 
different linguistic observations. All these assumptions, however, aim at modeling the optimal context information, 
be it syntactic and/or semantic, to perform the word prediction. The resulting models can be broadly classified into two main categories:
long and short range context models. The rest of this section presents a brief overview of these categories with a particular focus 
on Neural Network (NN)-based models.

\subsection{Short Range Context}
\label{ssec:SRC}
This category includes models that approximate~(\ref{eq:prob}) based on the Markov dependence assumption of 
order $N-1$. That is, the prediction of the current word depends only on the last $N-1$ words in the history.
In this case, (\ref{eq:prob}) becomes
%
\begin{equation}
  \label{eq:fnn-app-prob}
	 \displaystyle{ p(w_1^T) \approx \prod_{t=1}^T{p(w_t|w_{t-N+1}^{t-1})} }
\end{equation}
The most popular methods that subscribe in this category are the $N$-gram models~\cite{Rosenfeld2000,KN1995}
 as well as the FFNN model~\cite{Bengio2003}, which estimates each of the terms involved in this product, 
i.e, $p(w_t|w_{t-N+1}^{t-1})$ in a single bottom-up evaluation of the network.

Although these methods perform well and are easy to learn, the natural languages that they try to encode, however, 
are not generated under a Markov model due to their dynamic nature and the long range dependencies they manifest. 
Alleviating this assumption led to an extensive research to develop more suitable modeling techniques.

\subsection{Long Range Context}
\label{ssec:LRC}

\begin{figure}[h!]
\centering
\includegraphics[width=8cm]{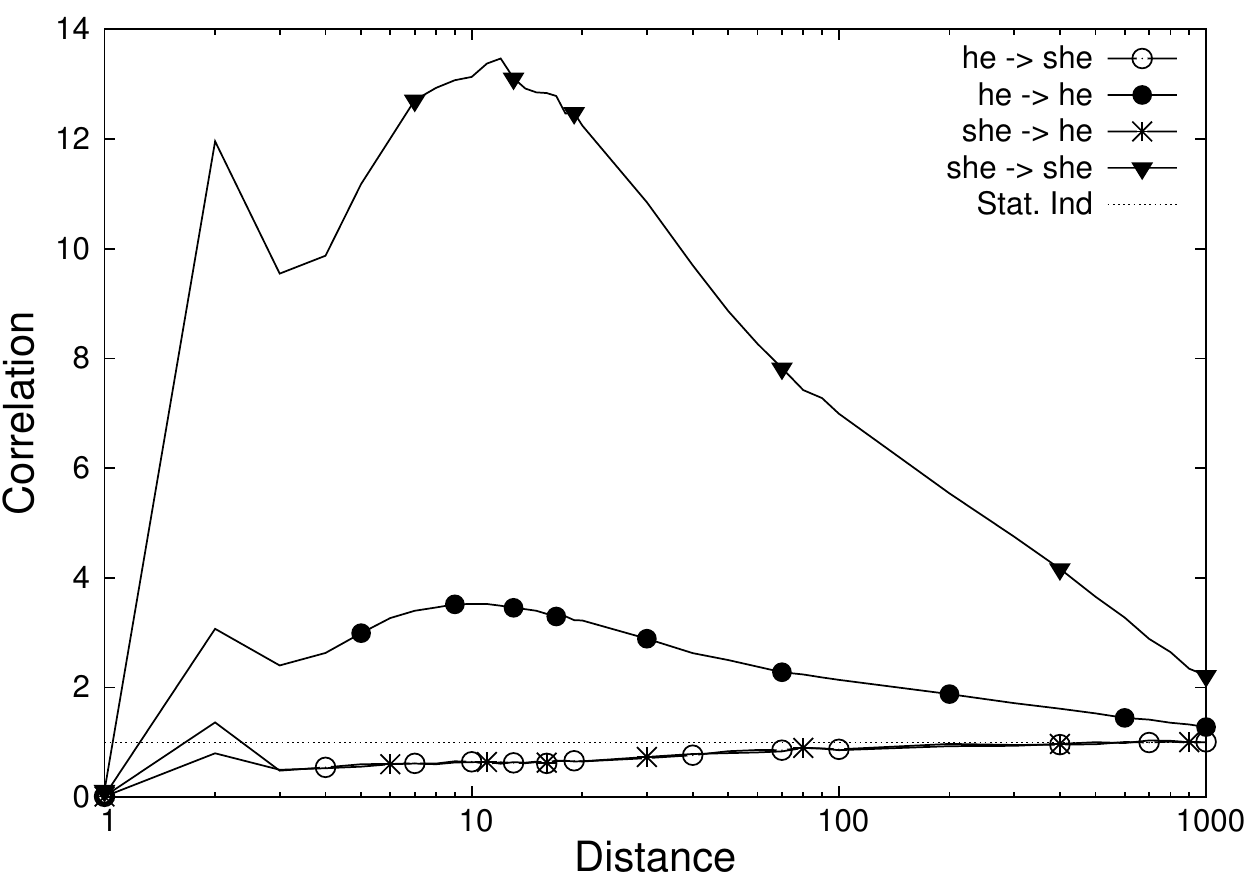}
\caption{Variation of word triggering correlations for pronouns over large distances.}
\label{fig:corr}
\end{figure}

Conventionally, N-gram related LMs have not been built to capture long linguistic dependencies, 
although significant word triggering information is still available for large contexts. 
To illustrate such triggering correlations spread over a large context, we use correlation defined over a distance $d$, 
given by $c_d(w_1,w_2) = \frac{P_d(w_1,w_2)}{P(w_1)P(w_2)}$. A value greater than 1 shows that it is more likely that the word $w_1$ 
follows $w_2$ at a distance $d$ than expected without the occurrence of $w_2$. In Figure~\ref{fig:corr}, we show the variation of this 
correlation for pronouns with the distance $d$. It can be observed that seeing another ``he" about twenty words after having seen a 
first ``he" is much more likely. A similar observation can be made for the word ``she". It is, however, surprising that 
seeing ``he" after ``he" is three times more likely than seeing ``she" after ``she", so ``he" is much more predictive. 
In the cases of cross-word triggering of ``he" $\rightarrow$ ``she" and ``she" $\rightarrow$ ``he", we find that the correlation 
is suppressed in comparison to the same word triggering for distances larger than three. In summary, Figure~\ref{fig:corr} 
demonstrates that word triggering information exists at large distances, even up to one thousand words. These conclusions
were confirmed by similar correlation experiments that we conducted for different types of words and triggering relations.

In order to model this long-term correlation and overcome the restrictive Markov assumption, 
recurrent language models have been proposed to approximate (\ref{eq:prob}) according to
\begin{equation}
  \label{eq:rnn-app-prob}
\!\! \displaystyle{ p(w_1^T) \approx \prod_{t=1}^T{p(w_t|w_{t-1}, h_{t-1})} = \prod_{t=1}^T{p(w_t|h_{t})} } \!\!\!\!\!\!
\vspace{-1mm}
\end{equation}

In NN-based recurrent models, $h_{t}$ is a context vector which represents the complete 
history, and modeled as a hidden state that evolves within the network.

\subsubsection{Elman-Type RNN-based LM}
\label{sssec:LRC}
\begin{figure}[h!]
	  \centering
\vspace{1mm}
	  \includegraphics[width=5cm,height=4cm]{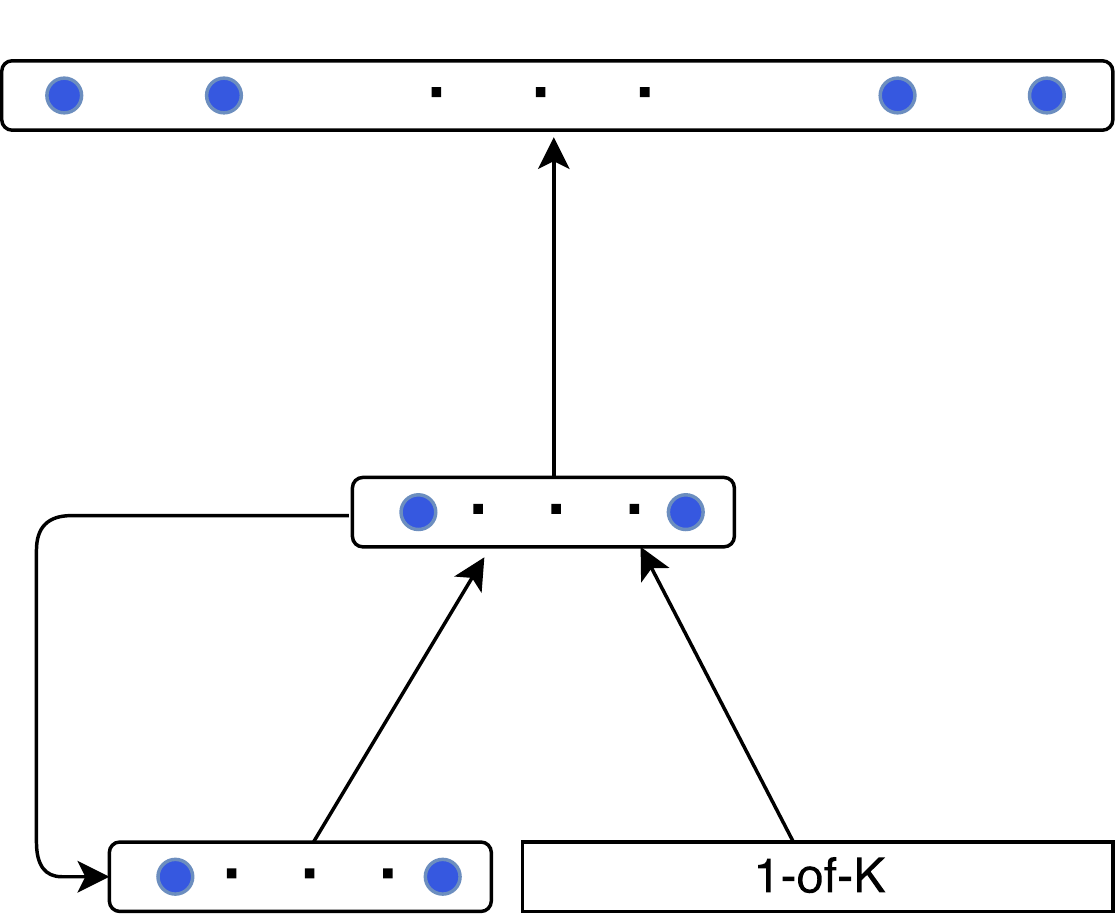}
	  \caption{Elman RNN architecture.}
	  \label{fig:rnn}
\end{figure}

The classical RNN~\cite{Mikolov2010} estimates each of the product terms in (\ref{eq:rnn-app-prob}) according to
%
\begin{align}
 \label{eqn:eqrnn-1}
  H_{t} &= f \left( X_{t-1} + V \cdot H_{t-1} \right) \\
   \label{eqn:eqrnn-2}
    P_t   &= g \left( W \cdot H_{t}\right)
 \end{align}

where $X_{t-1}$ is a continuous representation (i.e, embedding) of the 
word $w_{t-1}$, $V$ encodes the recurrent connection weights and $W$ is the hidden-to-output 
connection weights. These parameters define the network and are learned during training. 
Moreover, $f(\cdot)$ is an activation function, whereas $g(\cdot)$ is the softmax function. 
Figure~(\ref{fig:rnn}) shows an example of the standard RNN architecture. 

Theoretically, the recurrent connections of an RNN allow the context to indefinitely 
cycle in the network and thus, modeling long context. In practice, however, ~\newcite{HaiSon2012} 
have shown that this information changes quickly over time, and that it is experimentally 
equivalent to an 8-gram FFNN. This observation was confirmed by the experiments that we report in this paper.
 
\subsubsection{Long-Short Term Memory Network}
\label{sssec:LSTM}

\begin{figure}[!h]
  \centering
	\vspace{1mm}
   \includegraphics[totalheight=0.22\textheight, width=0.46\textwidth]{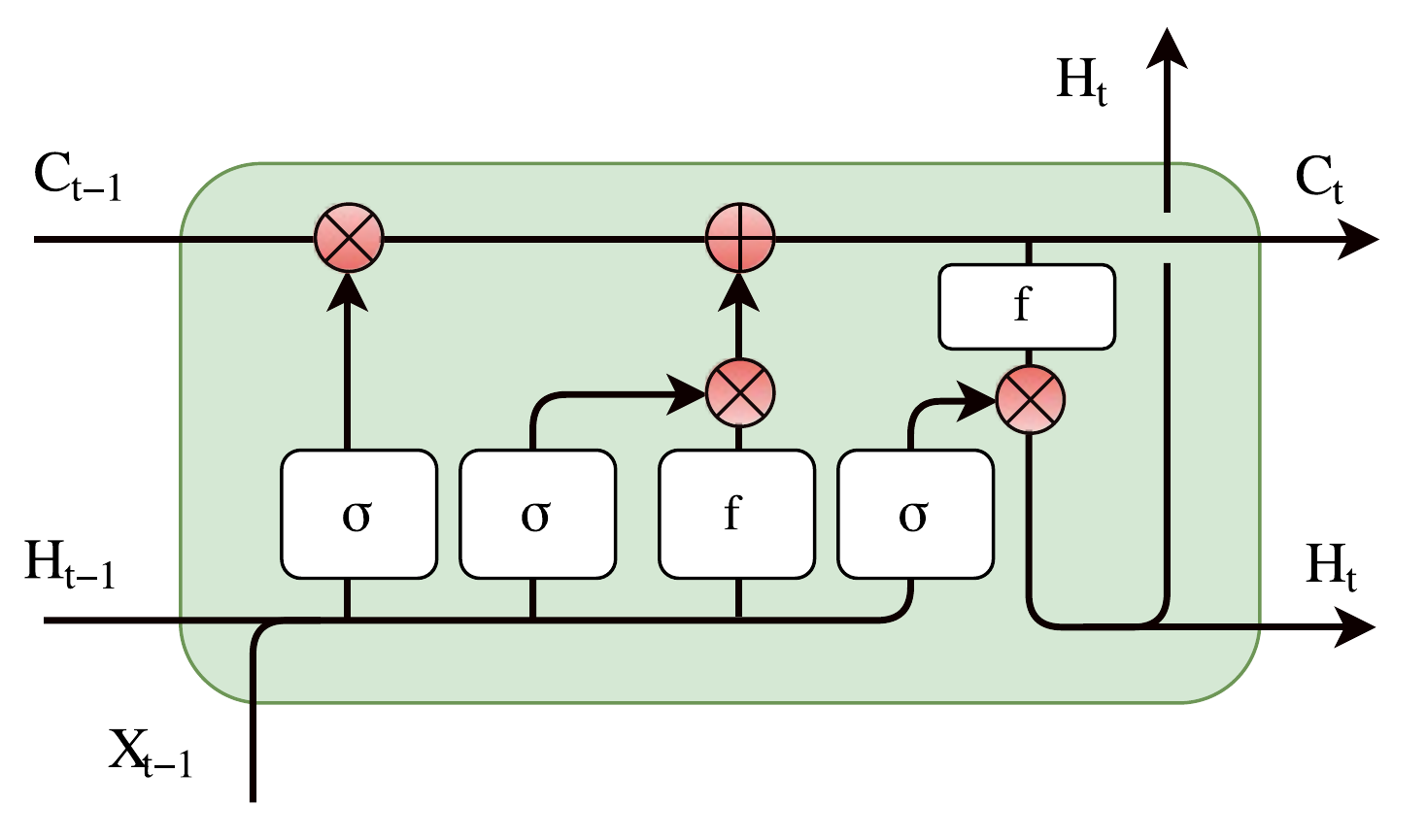}
  \caption{Block diagram of the recurrent module of an LSTM network.} 
  \label{fig:LSTM}
\end{figure}

In order to alleviate the rapidly changing context issue in standard RNNs and control the longevity of the dependencies modeling in the network, the LSTM architecture ~\cite{Sundermeyer12} introduces an internal memory state $C_{t}$, which explicitly controls the amount of information, to forget or to add to the network, before estimating the current hidden state. Formally, this is done according to
%
\begin{align}
 \label{eqn:eqlstm-1}
  {\lbrace i,f,o\rbrace}_t &= \sigma \left(  U^{i,f,o} \cdot X_{t-1} + V^{i,f,o} \cdot H_{t-1} \right) \\
  \label{eqn:eqlstm-2}
  \tilde{C_{t}} &= f \left( U^{c} \cdot X_{t-1} +  V^{c} \cdot H_{t-1} \right) \\
  \label{eqn:eqlstm-3}
  C_t &= f_t \odot C_{t-1} + i_t \odot \tilde{C_{t}}  \\
  \label{eqn:eqlstm-4}
  H_t &= o_t \odot f \left(C_t \right) \\
  \label{eqn:eqlstm-5}
    P_t &= g \left( W \cdot H_{t} \right)
 \end{align}
where $\odot$ is the element-wise multiplication operator, $\tilde{C_{t}}$ is the memory candidate, whereas $i_t,f_t$ and $o_t$ are the input, forget and output gates of the network, respectively. Figure~\ref{fig:LSTM} illustrates the recurrent module of an LSTM network. Learning of an LSTM model requires the training of the network parameters $U^{i,f,o,c}, V^{i,f,o,c}$ and $W$. 

Although LSTM models have been shown to outperform classical RNN in modeling long range dependencies, they do not explicitly model long/short context but rather use a single state to encode the global linguistic context. 

%

\section{Multi-Span Language Models}
\label{sec:MSLM}
The attempts to learn and combine short and long range dependencies in language modeling led to what is known as multi-span LMs~\cite{Bellegarda:1998}. The goal of these models is to learn the various constraints, both local and global, that are present in a language. This is typically done using two different models, which separately learn the local and global context, and then combine their resulting linguistic information to perform the word prediction. For instance,~\newcite{Bellegarda1998} proposed to use Latent Semantics Analysis (LSA) to capture the global context, and then combine it with the standard $N$-gram models, which capture the local context, whereas ~\newcite{Mikolov2012} proposed to model the global topic information using Latent Dirichlet Allocation (LDA), which is then combined with an RNN-based LM. This idea is not particular to language modeling but has been also used in other Natural Language Processing (NLP) tasks, e.g.,~\newcite{Tasos2014} proposed to use a local/global model to perform a spoken language understanding task. 

\subsection{Long-Short Range Context Network}
\label{sec:LSRC}
Following the line of thoughts in~\cite{Bellegarda1998,Mikolov2012}, we propose a new multi-span model, which takes advantage of the LSTM ability to model long range context while, simultaneously, learning and integrating the short context through an additional recurrent, local state. In doing so, the resulting Long-Short Range Context (LSRC) network is able to separately model the short/long context while it dynamically combines them to perform the next word prediction task. Formally, this new model is defined as
%
\begin{align}
  \label{eqn:eqlsrc-1}
   H_{t}^{l} &= f \left( X_{t-1} +  U^{c}_l  \cdot H_{t-1}^{l}\right) \\
  \label{eqn:eqlsrc-2}
  {\lbrace i,f,o\rbrace}_t &= \sigma \left(  V^{i,f,o}_l \cdot H_{t}^{l} + V^{i,f,o}_g \cdot  H_{t-1}^g \right) \\
  \label{eqn:eqlsrc-3}
  \tilde{C_{t}} &= f \left(V^{c}_l \cdot H_{t}^{l} + V^{c}_g \cdot H_{t-1}^g \right) \\
  \label{eqn:eqlsrc-4}
  C_t &= f_t \odot C_{t-1} + i_t \odot \tilde{C_{t}}  \\
  \label{eqn:eqlsrc-5}
  H_t^g &= o_t \odot f \left(C_t \right) \\
  \label{eqn:eqlsrc-6}
    P_t &= g \left( W \cdot H_{t}^g \right)
 \end{align}

Learning of an LSRC model requires the training of the local parameters $V^{i,f,o,c}_l$ and $U_l^c$, the global parameters $V^{i,f,o,c}_g$ and the hidden-to-output connection weights $W$. This can be done using the standard Back-Propagation Through Time (BPTT) algorithm, which is typically used to train recurrent networks. 

\begin{figure}[!h]
  \centering
	\vspace{1mm}
   \includegraphics[totalheight=0.24\textheight, width=0.48\textwidth]{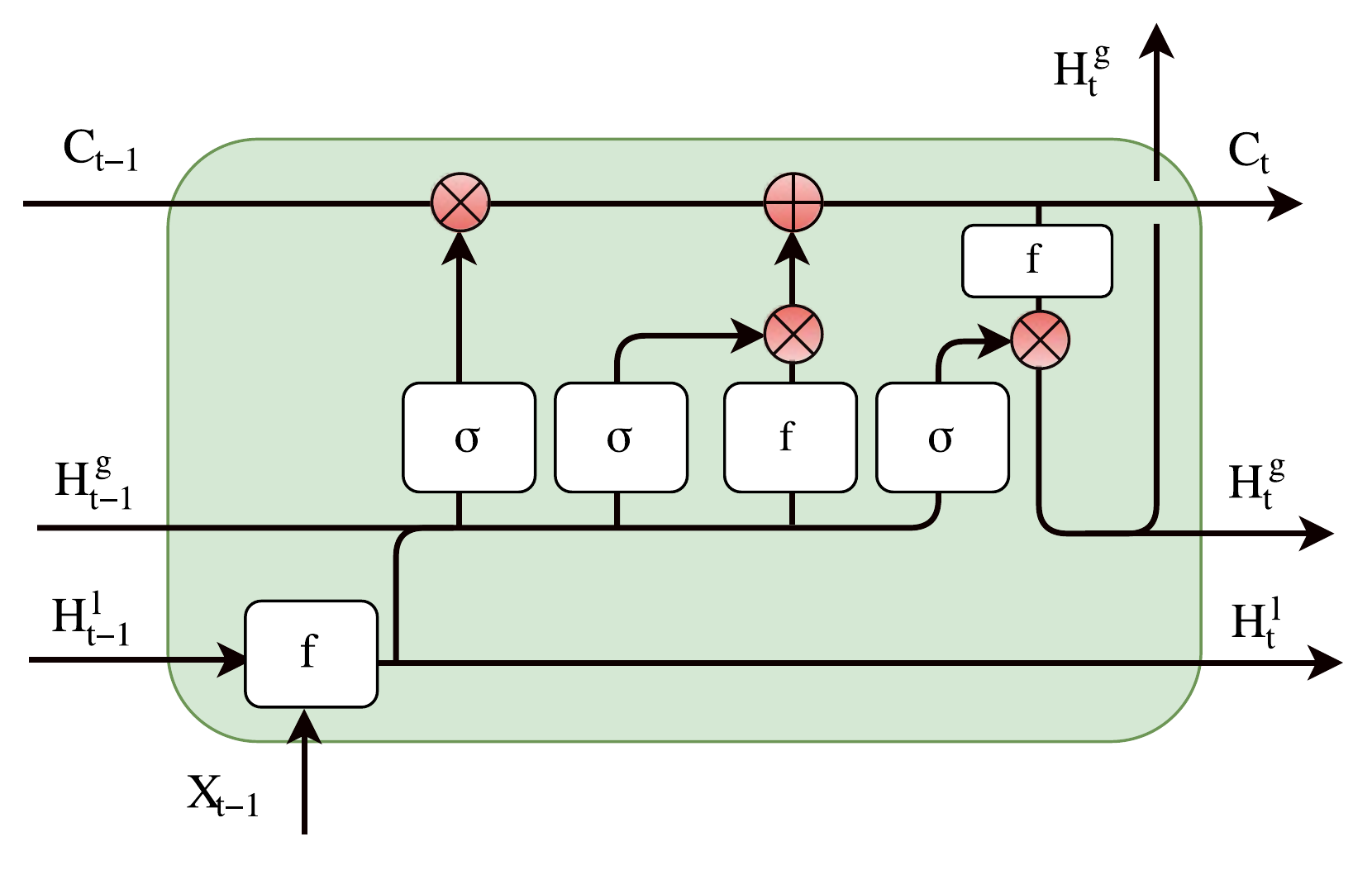}
  \caption{Block diagram of the recurrent module of an LSRC network.} 
  \label{fig:LSRC}
\end{figure}

The proposed approach uses two hidden states, namely, $H_t^l$ and $H_t^g$ to model short and long range context, respectively. More particularly, the local state $H_t^l$ evolves according to (\ref{eqn:eqlsrc-1}) which is nothing but a simple recurrent model as it is defined in (\ref{eqn:eqrnn-1}). In doing so, $H_t^l$ is expected to have a similar behavior to RNN, which has been shown to capture local/short context (up to 10 words), whereas the global state $H_t^g$ follows the LSTM model, which is known to capture longer dependencies (see example in Figure~\ref{fig:TempCorr}). The main difference here, however, is the dependence of the network modules (gates and memory candidate) on the previous local state $H_t^l$ instead of the last seen word $X_{t-1}$. This model is based on the assumption that the local context carries more linguistic information, and is therefore, more suitable to combine with the global context and update LSTM, compared to the last seen word. Figure~\ref{fig:LSRC} illustrates the recurrent module of an LSRC network. It is worth mentioning that this model was not particularly developed to separately learn syntactic and semantic information. This may come, however, as a result of the inherent local and global nature of these two types of linguistic properties.

\subsection{Context Range Estimation}
\label{sec:LSRCE}

For many NLP applications, capturing the global context information can be a crucial component to develop successful systems. This is 
mainly due to the inherent nature of languages, where a single idea or topic can span over few sentences, paragraphs or a complete document.
LSA-like approaches take advantage of this property, and aim at extracting some hidden ``concepts'' that best explain the data in a low-dimension ``semantic space''. 
To some extent, the hidden layer of LSRC/LSTM can be seen as a vector in a similar space. The information stored in this vector, however, changes
continuously based on the processed words. Moreover, interpreting its content is generally difficult. As an alternative, measuring the 
temporal correlation of this hidden vector can be used as an indicator of the ability of the network to model short and long context dependencies.
Formally, the temporal correlation of a hidden state $H$ over a distance $d$ is given by
\begin{equation}
 c_d = \frac{1}{D} \sum_{t=1}^{t=D} {SM(H_t,H_{t+d})}
\end{equation}
where $D$ is the test data size in words and $SM$ is a similarity measure such as the \textit{cosine similarity}. 
This measure allows us to evaluate how fast does the information stored in the hidden state change over time. 

\begin{figure}[!h]
  \centering
	   \vspace{1mm}
		 \hspace{-3mm}
   \includegraphics[trim=5 0 15 0 ,clip, totalheight=0.19\textheight, width=0.485\textwidth]{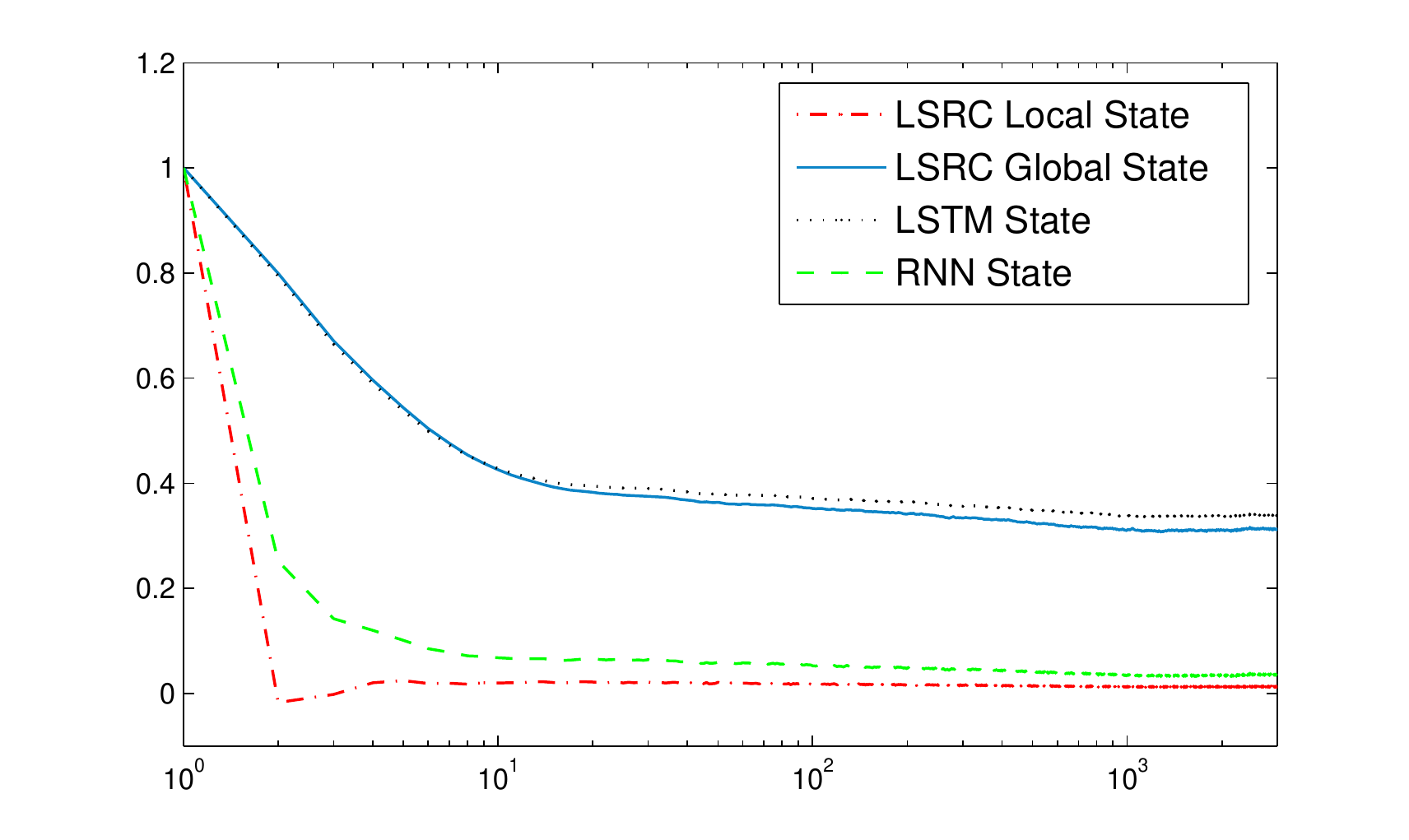} 
  \caption{Temporal correlation of the proposed network in comparison to LSTM and RNN.} 
  \label{fig:TempCorr}
\end{figure}

In Figure~\ref{fig:TempCorr}, we show the variation of this temporal correlation for the local and global states of 
the proposed LSRC network in comparison to RNN and LSTM for various values of the distance $d$ (up to 3000). 
This figure was obtained on the test set of the Penn Treebank (PTB) corpus, described in Section~(\ref{sec:EXP}). 
The main conclusion we can draw from this figure is the ability of the LSRC local and global states (trained jointly) 
to behave in a similar fashion to RNN and LSTM states (trained separately), respectively. We can also conclude 
that the LSRC global state and LSTM are able to capture long range correlations, whereas the 
context changes rapidly over time in RNN and LSRC local state.

\section{Experiments and Results}
\label{sec:EXP}
\subsection{Experimental Setup}
We evaluated the proposed architecture on two different benchmark tasks. 
The first set of experiments was conducted on the commonly used 
Penn Treebank (PTB) corpus using the same experimental setup adopted in~\cite{Mikolov2011} and~\cite{FOFE2015}. 
Namely, sections 0-20 are used for training while sections 21-22 and 23-24 are used for validation an testing,
respectively. The vocabulary was limited to the most 10k frequent words while the remaining words were mapped to the token $<$unk$>$. 

In order to evaluate how the proposed approach performs on large corpora in comparison to other methods, 
we run a second set of experiments on the Large Text Compression Benchmark (LTCB)~\cite{Mahoney2011}. 
This corpus is based on the enwik9 dataset which contains the first $10^9$ bytes of enwiki-20060303-pages-articles.xml. 
We adopted the same training-test-validation data split as well as the the same data 
processing\footnote{\label{FOFE}All the data processing steps described here for PTB 
and LTCB were performed using the FOFE toolkit in~\cite{FOFE2015}, 
which is available at \url{https://wiki.eecs.yorku.ca/lab/MLL/_media/projects:fofe:fofe-code.zip}} which were used in~\cite{FOFE2015}. 
The vocabulary is limited to the most 80k frequent words with all remaining words replaced by $<$unk$>$. 
Details about the sizes of these two corpora can be found in Table~\ref{tab:corpora}. 
\begin{table}[!th]
  \centerline{
  \begin{tabular}{| c || c |c | c |}
    \hline
    Corpus  & Train & Dev & Test \\
    \hline
    \hline
    PTB & 930K & 74K & 82K    \\
    \hline
    LTCB & 133M & 7.8M & 7.9M \\
    \hline
  \end{tabular}}
    \caption{\label{tab:corpora} Corpus size in number of words.}
\end{table}

Similarly to the RNN LM toolkit\footnote{\label{RNNLM}The RNN LM toolkit is available at \url{http://www.rnnlm.org/}}~\cite{Mikolov2011}, 
we have used a single end sentence tag between each two consecutive sentences, whereas the begin sentence tag was not 
included\footnote{This explains the difference in the corpus size compared to the one reported in~\cite{FOFE2015}.}.
\subsection{Baseline Models}
The proposed LSRC architecture is compared to different LM approaches that model short or long range context. 
These include the commonly used $N$-gram Kneser-Ney (KN)~\cite{KN1995} model with and without cache~\cite{Kuhn1990},
as well as different feedforward and recurrent neural architectures. 
For short (fixed) size context models, we compare our method to 1) the FFNN-based LM~\cite{Bengio2003}, as well as 2) the Fixed-size 
Ordinally Forgetting Encoding (FOFE) approach, which is implemented in~\cite{FOFE2015} as a sentence-based model. 
For these short size context models, we report the results of different history window sizes (1, 2 and 4). 
The $1^{st}$, $2^{nd}$ and $4^{th}$-order FOFE results were either reported in~\cite{FOFE2015} or obtained using 
the freely available FOFE toolkit~\footref{FOFE}. 

For recurrent models that were designed to capture long term context, we compared the proposed approach to 3) the full RNN 
(without classes)~\cite{Mikolov2011}, 4) to a deep RNN (D-RNN)\footnote{\label{DRNN}The deep RNN results 
were obtained using $L_p$ and maxout units, dropout regularization and gradient control techniques, which 
are known to significantly improve the performance. None of these techniques, however, were used in our 
experiments.}~\cite{Pascanu2013}, which investigates different approaches to construct mutli-layer RNNs, 
and finally 5) to the LSTM model~\cite{Sundermeyer12}, which explicitly regulates the amount of information that propagates in the network. 
The recurrent models results are reported for different numbers of hidden layers (1 or 2).  
In order to investigate the impact of deep models on the LSRC architecture, we added a single hidden, non-recurrent layer (of size 400 for PTB
and 600 for the LTCB experiments) to the LSRC model (D-LSRC). 
This was sufficient to improve the performance with a negligible increase in the number of model parameters. 

\subsection{PTB Experiments}

For the PTB experiments, the FFNN and FOFE models use a word embedding size of 200, whereas the hidden layer(s) size is fixed at 400,
with all hidden units using the Rectified Linear Unit (ReLu) i.e., $f(x)=max(0,x)$
as activation function. We also use the same learning setup adopted in~\cite{FOFE2015}. Namely, 
we use the stochastic gradient descent algorithm with a mini-batch size of 200, 
the learning rate is initialized to 0.4, the momentum is set to 0.9, the weight 
decay is fixed at $4\times10^{-5}$, whereas the training is done in epochs. 
The weights initialization follows the normalized initialization proposed 
in~\cite{Glorot2010}. Similarly to~\cite{Mikolov2010}, 
the learning rate is halved when no significant improvement of the validation data log-likelihood is
observed. Then, we continue with seven more epochs while halving the learning rate after each epoch. 

Regarding the recurrent models, we use $f=tanh(\cdot)$ as activation function for all recurrent 
layers, whereas "$f=sigmoid(\cdot)$" is used for the input, forget and output gates of LSTM and LSRC. 
The additional non-recurrent layer in D-LSRC, however, uses the ReLu activation function. The word embedding 
size was set to 200 for LSTM and LSRC whereas it is the same as the hidden layer size for RNN 
(result of the RNN equation~\ref{eqn:eqrnn-1}). In order to illustrate the effectiveness of the LSRC model,
we also report the results when the embedding size is fixed at 100, LSRC(100). The training uses the BPTT algorithm for 
5 time steps. Similarly to short context models, the mini-batch was set to 200. The learning rate, 
however, was set to 1.0 and the weight decay to $5\times10^{-5}$. The use of momentum did not lead to any additional improvement.  
Moreover, the data is processed sequentially without any sentence independence assumption. Thus, the recurrent models 
will be able to capture long range dependencies that exist beyond the sentence boundary.

In order to compare the model sizes, we also report the Number of Parameters (NoP) to train for each of the models above.

\renewcommand{\tabcolsep}{2pt}
\begin{table}[!h]
  \centerline{
  \begin{tabular}{| c | c | c | c || c | c | c || c |}
    \hhline{~|-|-|-||-|-|-||-|}
	 	\hhline{~|-|-|-||-|-|-||-|}
     \multicolumn{1}{c|}{} & \multicolumn{3}{|>{\columncolor{grey}}c||}{ model } &  
		\multicolumn{3}{|>{\columncolor{grey}}c||}{ model+KN5 } & \multicolumn{1}{|>{\columncolor{grey}}c|}{ NoP } \\
     \hline
      N-1= & 1 & 2 & 4 & 1 & 2 & 4 & 4  \\
      \hline
			KN        & 186 & 148 & 141  & --- & --- & --- & --- \\
			KN+cache  & 168 & 134 & 129  & --- & --- & --- & --- \\
			\hline
			\hhline{~|-|-|-||-|-|-||-|}
      & \multicolumn{7}{|>{\columncolor{grey2}}c|}{ 1 Hidden Layer }  \\
      \hline
      FFNN       & 176 & 131 & 119  & 132 & 116 & 107 & 6.32M \\
      FOFE      & 123 & 111 & 112  & 108 & 100 & 101 & 6.32M \\
      \hline
      \hhline{~|-|-|-||-|-|-||-|}
      & \multicolumn{7}{|>{\columncolor{grey2}}c|}{ Recurrent Models (1 Layer) }  \\
      \hline
      RNN & \multicolumn{3}{c||}{ 117 } & \multicolumn{3}{c||}{ 104 } & 8.16M \\
      \hline
			LSTM (1L) & \multicolumn{3}{c||}{ 113 } & \multicolumn{3}{c||}{ 99 } & 6.96M \\
      \hline
      LSRC(100) & \multicolumn{3}{c||}{ 109 } & \multicolumn{3}{c||}{ 96 } & 5.81M \\
			\hline
			LSRC(200) & \multicolumn{3}{c||}{ 104 } & \multicolumn{3}{c||}{ 94 } & 7.0M \\
      \hline
     \hhline{~|-|-|-||-|-|-||-|}
      & \multicolumn{7}{|>{\columncolor{grey2}}c|}{ 2 Hidden Layers }  \\
      \hline
      FFNN       & 176 & 129 & 114  & 132  & 114 & 102 & 6.96M \\
      FOFE      & 116 & 108 & 109  & 104  & 98  &  97 & 6.96M \\
      \hline 
     \hhline{~|-|-|-||-|-|-||-|}
      & \multicolumn{7}{|>{\columncolor{grey2}}c|}{ Deep Recurrent Models }  \\
      \hline
      D-LSTM (2L)  & \multicolumn{3}{c||}{ 110 } & \multicolumn{3}{c||}{ 97 } & 8.42M \\
      \hline
      D-RNN\footref{DRNN} (3L) & \multicolumn{3}{c||}{ 107.5 } & \multicolumn{3}{c||}{ NR } & 6.16M \\
      \hline
      D-LSRC(100)  & \multicolumn{3}{c||}{ 103 } & \multicolumn{3}{c||}{ 93 } & 5.97M \\
      \hline
			D-LSRC(200)  & \multicolumn{3}{c||}{ 102 } & \multicolumn{3}{c||}{ 92 } & 7.16M \\
      \hline
  \end{tabular} }
    \caption{\label{tab:ptb} LMs performance on the PTB test set.}
\end{table}

Table~\ref{tab:ptb} shows the perplexity evaluation on the PTB test set. As a first observation, we can clearly see that
the proposed approach outperforms all other models for all configurations, in particular, RNN and LSTM. 
This observation includes other models that were reported in the literature, such as random forest LM~\cite{Xu2007}, 
structured LM~\cite{Filimonov2009} and syntactic neural network LM~\cite{Emami2004}. More particularly, 
we can conclude that LSRC, with an embedding size of 100, achieves a better performance than all other models while
reducing the number of parameters by $\approx29\%$ and $\approx17\%$ compared to RNN and LSTM, respectively.  
Increasing the embedding size to 200, which is used by the other models, improves significantly the performance
with a resulting NoP comparable to LSTM.  The significance of the improvements obtained here over LSTM
were confirmed through a statistical significance t-test, which led to p-values $\leq 10^{-10}$ for a significance level
of $5\%$ and $0.01\%$, respectively.

The results of the deep models in Table~\ref{tab:ptb} also show that adding a single non-recurrent hidden layer to LSRC can 
significantly improve the performance. In fact, the additional layer bridges the gap between the LSRC models with an embedding size 
of 100 and 200, respectively. The resulting architectures outperform the other deep recurrent 
models with a significant reduction of the number of parameters (for the embedding size 100), 
and without usage of dropout regularization, $L_p$ and maxout 
units or gradient control techniques compared to the deep RNN\footref{DRNN}(D-RNN).  

We can conclude from these experiments that the explicit modeling of short and long range dependencies using two separate hidden 
states improves the performance while significantly reducing the number of parameters. 
%
%
\begin{figure}[h!]
	  \centering
	   \hspace{-2mm}
	  \includegraphics[trim=0 0 10 0 ,clip, width=7.9cm,height=5cm]{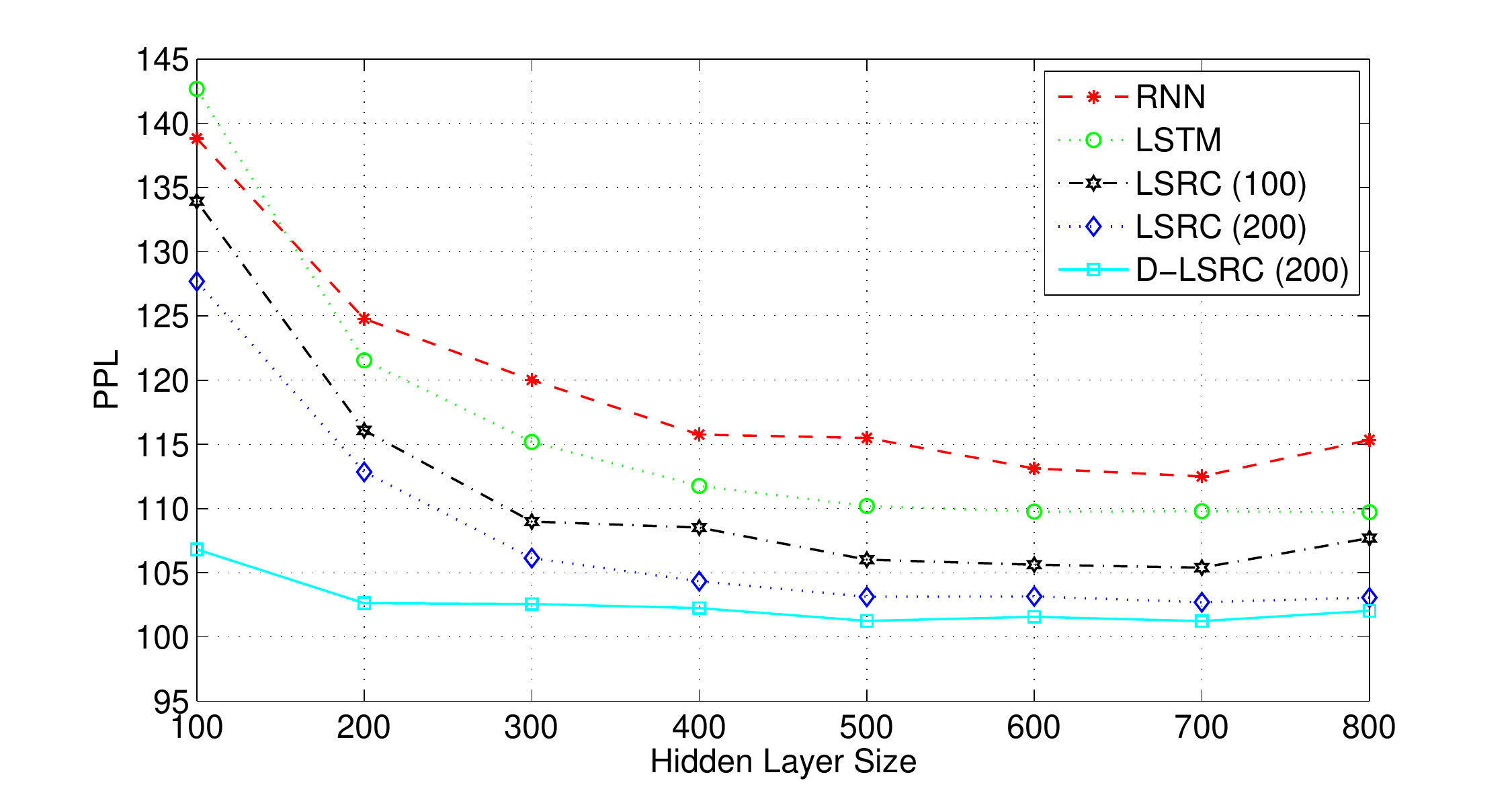}
	  \caption{Perplexity of the different NN-based LMs with different hidden layer sizes on the PTB test set.}
	  \label{fig:ppl}
\end{figure}

In order to show the consistency of the LSRC improvement over the other recurrent models, 
we report the variation of the models performance with respect to the hidden layer size in 
Figure~\ref{fig:ppl}. This figure shows that increasing the LSTM or RNN hidden layer size 
could not achieve a similar performance to the one obtained using LSRC with a small layer 
size (e.g., 300). It is also worth mentioning that this observation holds when comparing a 
2-recurrent layers LSTM to LSRC with an additional non-recurrent layer.

\subsection{LTCB Experiments}
			
The LTCB experiments use the same PTB setup with minor modifications. 
The results shown in Table~\ref{tab:ltcb} follow the same experimental setup proposed in~\cite{FOFE2015}. 
More precisely, these results were obtained without use of momentum or weight decay (due to the 
long training time required for this corpus), the mini-batch size was set to 400, 
the learning rate was set to 0.4 and the BPTT step was fixed at 5.  
The FFNN and FOFE architectures use 2 hidden layers of size 600, whereas RNN, LSTM 
and LSRC have a single hidden layer of size 600. Moreover, the word embedding size was set to 
200 for all models except RNN, which was set to 600. We also report results for an LSTM with 2 recurrent layers as well as for 
LSRC with an additional non-recurrent layer. The recurrent layers are marked with an ``R'' in Table~\ref{tab:ltcb}.
\renewcommand{\tabcolsep}{1.5pt}
\begin{table}[!th]
\centerline{
\begin{tabular}{| c | c | c | c || c |}
\hhline{~|-|-|-||-|}
\hhline{~|-|-|-||-|} 
\multicolumn{1}{c|}{} & \multicolumn{3}{|>{\columncolor{grey}}c||}{model} & \multicolumn{1}{|>{\columncolor{grey}}c|}{NoP} \\
\hline
Context Size M=N-1 & 1 & 2 & 4 & 4  \\
\hline
KN       &  239  &  156  & 132 & ---  \\
KN+cache &  188  &  127  & 109  & ---  \\
\hline
FFNN  [M*200]-600-600-80k & 235 & 150 & 114 & 64.84M \\
FOFE [M*200]-600-600-80k & 112 & 107 & 100 & 64.84M \\
\hline
\hline
RNN [600]-R600-80k     & \multicolumn{3}{c||}{ 85 } & 96.36M   \\
\hline
\hline
LSTM [200]-R600-80k      & \multicolumn{3}{c||}{ 66 } &  65.92M \\
LSTM [200]-R600-R600-80k & \multicolumn{3}{c||}{ 61 } & 68.80M \\
\hline
\hline
LSRC [200]-R600-80k     & \multicolumn{3}{c||}{ 63 } &  65.96M \\
LSRC [200]-R600-600-80k & \multicolumn{3}{c||}{ 59 } & 66.32M \\
\hline
\end{tabular} }
\caption{\label{tab:ltcb} LMs performance on the LTCB test set.}
\end{table}

The results shown in Table~\ref{tab:ltcb} generally confirm the conclusions we drew from the 
PTB experiments above. In particular, we can see that the proposed LSRC model largely outperforms 
all other models. In particular, LSRC clearly outperforms LSTM with a negligible 
increase in the number of parameters (resulting from the additional $200\times200=0.04M$ 
local connection weights $U^c_l$) for the single layer results. We can also see that this 
improvement is maintained for deep models (2 hidden layers), where the LSRC model achieves 
a slightly better performance while reducing the number of parameters by $\approx2.5M$
and speeding up the training time by $\approx20\%$ compared to deep LSTM.

The PTB and LTCB results clearly highlight the importance of recurrent models to capture long 
range dependencies for LM tasks. The training of these models, however, requires large amounts of 
data to significantly outperform short context models. This can be seen in the performance 
of RNN and LSTM in the PTB and LTCB tables above. We can also conclude from these results that the 
explicit modeling of long and short context in a multi-span model can lead to a 
significant improvement over state-of-the are models.

\section{Conclusion and Future Work}
\label{sec:CC}
We investigated in this paper the importance, followed by the ability, of standard neural networks to
encode long and short range dependencies for language modeling tasks. We also showed that these models
were not particularly designed to, explicitly and separately, capture these two linguistic information. 
As an alternative solution, we proposed a novel long-short range context network, which takes advantage of the LSTM
ability to capture long range dependencies, and combines it with a classical RNN network, which typically 
encodes a much shorter range of context. In doing so, this network is able to encode the short 
and long range linguistic dependencies using two separate network states that evolve in time.
Experiments conducted on the PTB and the large LTCB corpus have shown that
the proposed approach significantly outperforms different state-of-the are neural network 
architectures, including LSTM and RNN, even when smaller architectures are used.  
This work, however, did not investigate the nature of the long 
and short context encoded by this network or its 
possible applications for other NLP tasks. This is part of our future work.

\section*{Acknowledgments}
This work was in part supported by the Cluster of Excellence for Multimodal Computing and Interaction, 
the German Research Foundation (DFG) as part of SFB 1102, the EU FP7 Metalogue project 
(grant agreement number: 611073) and the EU Malorca project (grant agreement number: 698824).

\bibliography{emnlp2016}
\bibliographystyle{emnlp2016}

\end{document}